\documentclass[journal]{IEEEtran}

\usepackage[T1]{fontenc}
\usepackage[utf8]{inputenc}
\usepackage{cite}
\usepackage{csquotes}
\usepackage{fancyvrb}
\usepackage[nolist]{acronym}
\usepackage{pifont}
\usepackage[dvipsnames]{xcolor}
\usepackage{colortbl}
\usepackage{graphicx}
\usepackage[caption=false,font=footnotesize]{subfig}
\usepackage{float}
\usepackage{tabularx}
\usepackage{makecell}
\usepackage{array}
\usepackage{multirow}
\usepackage{booktabs}
\usepackage{dblfloatfix}
\usepackage{placeins}
\usepackage{amsmath}
\usepackage{threeparttable}
\usepackage{afterpage}
\usepackage{wrapfig}
\usepackage{tikz}
\usetikzlibrary{arrows.meta,positioning,fit,shapes.geometric,calc}
\usepackage{helvet}
\usepackage[hyphens,spaces,obeyspaces]{url}

\renewcommand{\arraystretch}{0.8}
\newcolumntype{C}{>{\centering\arraybackslash}X}
\newcommand{\diagramfont}{\sffamily\scriptsize\bfseries\boldmath}
\newlength{\diagramgap}
\newlength{\pipelinecolumnwidth}
\newlength{\pipelinecolumnstep}
\tikzset{
    diagram block/.style={
        rounded corners=1mm, align=center,
        minimum height=9mm,
        inner xsep=1mm, inner ysep=1mm,
        outer sep=0.3pt, font=\diagramfont, line width=0.6pt
    },
    diagram arrow/.style={
        -{Latex[length=1.8mm]}, line width=0.6pt, draw=black
    },
    diagram dashed arrow/.style={diagram arrow, dashed, draw=IEEEgray},
    diagram group/.style={
        dashed, rounded corners=1mm, line width=0.6pt, inner sep=1mm
    }
}
\definecolor{IEEEpurple}{HTML}{981D97}
\definecolor{IEEEdpurple}{HTML}{772583}
\definecolor{IEEEcyan}{HTML}{00B5E2}
\definecolor{IEEEblue}{HTML}{00629B}
\definecolor{IEEEteal}{HTML}{009CA6}
\definecolor{IEEEdteal}{HTML}{007377}
\definecolor{IEEEgreen}{HTML}{78BE20}
\definecolor{IEEEolive}{HTML}{658D1B}
\definecolor{IEEEgray}{HTML}{75787B}
\definecolor{IEEElgray}{HTML}{F2F2F2}
\definecolor{IEEEred}{HTML}{BA0C2F}
\definecolor{TableHeader}{HTML}{E7EFF6}
\definecolor{TableHighlight}{HTML}{EDF7F2}

\newcommand{\tablehead}{\rowcolor{TableHeader}}
\newcommand{\keyresult}{\rowcolor{TableHighlight}}
\newcommand{\familyrule}{\addlinespace[1pt]\midrule\addlinespace[1pt]}

\begin{document}
\begin{acronym}
    \acro{AI}{artificial intelligence}
    \acro{API}{application programming interface}
    \acro{AP}{average precision}
    \acro{BAFE}{box-aligned feature extraction}
    \acro{BiLSTM}{bidirectional long short-term memory}
    \acro{CNN}{convolutional neural network}
    \acro{CNN-MV}{convolutional neural network for motion-vector propagation}
    \acro{CoViAR}{compressed video action recognition}
    \acro{CPU}{central processing unit}
    \acro{CUDA}{Compute Unified Device Architecture}
    \acro{CV}{computer vision}
    \acro{DCT}{discrete cosine transform}
    \acro{DFF}{deep feature flow}
    \acro{EMC}{external memory controller}
    \acro{F1}{harmonic mean of precision and recall}
    \acro{FGFA}{flow-guided feature aggregation}
    \acro{FLOP}{floating-point operation}
    \acro{FN}{false negative}
    \acro{FP}{false positive}
    \acro{FP16}{16-bit floating-point precision}
    \acro{FP32}{32-bit floating-point precision}
    \acro{FPS}{frames per second}
    \acro{GOP}{group of pictures}
    \acro{GPU}{graphics processing unit}
    \acro{ID}{identifier}
    \acro{IEEE}{Institute of Electrical and Electronics Engineers}
    \acro{ILSVRC}{ImageNet Large Scale Visual Recognition Challenge}
    \acro{IoU}{intersection over union}
    \acro{IRLS}{iteratively reweighted least squares}
    \acro{LSTM}{long short-term memory}
    \acro{LSTS}{learnable spatio-temporal sampling}
    \acro{MAC}{multiply-accumulate operation}
    \acro{mAP}{mean average precision}
    \acro{MEGA}{memory enhanced global-local aggregation}
    \acro{MOT}{multi-object tracking}
    \acro{MP4}{MPEG-4 Part 14 multimedia container}
    \acro{MPEG}{Moving Picture Experts Group}
    \acro{MV}{motion vector}
    \acro{MVP}{motion-vector propagation}
    \acro{DLA}{deep learning accelerator}
    \acro{NMS}{non-maximum suppression}
    \acro{NPU}{neural processing unit}
    \acro{ONNX}{Open Neural Network Exchange}
    \acro{OS}{operating system}
    \acro{P95}{95th percentile}
    \acro{P99}{99th percentile}
    \acro{PSLA}{progressive sparse local attention}
    \acro{RANSAC}{random sample consensus}
    \acro{RESPIRE}{reducing spatial-temporal redundancy for efficient edge-based industrial video analytics}
    \acro{ReLU}{rectified linear unit}
    \acro{RGB}{red, green, and blue}
    \acro{RMS}{root mean square}
    \acro{ROI}{region of interest}
    \acro{SELSA}{sequence level semantics aggregation}
    \acro{SoC}{system on chip}
    \acro{TII}{IEEE Transactions on Industrial Informatics}
    \acro{TP}{true positive}
    \acro{TPU}{tensor processing unit}
    \acro{UA-DETRAC}{University at Albany Detection and Tracking benchmark}
    \acro{VID}{video object detection}
    \acro{YOLO}{You Only Look Once}
\end{acronym}

\title{Analytical and Convolutional Neural Network-Based Motion-Vector Propagation for Efficient Video Object Detection}

\author{Ashiyana~Abdul Majeed,~\IEEEmembership{Member,~IEEE,},
        Mahmoud~Meribout,~\IEEEmembership{Senior Member,~IEEE,}
        and~Neethu~Joseph % <-this % stops a space
\thanks{Ashiyana Abdul Majeed, Dr Mahmoud Meribout, and Neethu Joseph are with the Department
of Computer and Information Engineering, Khalifa University, Abu Dhabi, UAE (email: 100059454@ku.ac.ae, mahmoud.meribout@ku.ac.ae, 100069410@ku.ac.ae).}}

\maketitle

\begin{abstract}
Continuous video analytics requires accurate localization at low latency within embedded power budgets. This paper presents a hardware-software design methodology that reuses codec \acp{MV} between detector invocations. Two alternative models support translation and scale changes: analytical motion-vector propagation (Analytical-MV) and learned propagation using a \ac{CNN} (CNN-MV). The learned model uses convolutional operations and independent object updates suited to parallel execution on an edge \ac{GPU}. Analytical-MV combines a harmonic-mean precision-recall score (F1) of 0.909 with a mean end-to-end latency of 9.03~ms and an energy of 0.177~J per frame, yielding the lowest latency and energy among the evaluated configurations. Relative to detection on every frame, it reduces mean latency by 25.9\% and energy per frame by 36.4\%. CNN-MV offers a different trade-off: its fastest configuration raises recall from Analytical-MV's 0.871 to 0.890, and lowers mean power from 19.64 to 17.32~W, at 18.42~ms latency and 0.319~J/frame. It is therefore useful when recall or operating power matters more than minimum latency and energy. Execution on a \ac{DLA} further reduces time-averaged GPU utilization relative to GPU execution. Host-processing optimization substantially improves both latency and energy, demonstrating the value of jointly designing temporal models and their execution pipelines.
\end{abstract}
\acresetall


\IEEEpeerreviewmaketitle

\section{Introduction}
\label{sec:introduction}

Continuous video object detection is an important component of intelligent traffic-monitoring infrastructure, where roadside edge devices must localize vehicles under strict latency, energy, and thermal constraints. Similar constraints arise in physical-AI systems such as autonomous vehicles, inspection drones, and mobile robots, which continuously process video while sharing limited computational resources with control, planning, and communication workloads. These tasks require frame-level localization at video rate within the compute, power, and thermal limits of compact edge devices. Processing every frame independently repeats substantial work because neighboring frames often contain the same objects. The resulting costs extend beyond neural inference to preprocessing and data movement, making complete-pipeline efficiency essential for sustained operation.

Temporal reuse offers a way to reduce this repeated computation. Video codecs already encode \acp{MV} for inter-frame prediction~\cite{ituH264}, and compressed-domain learning has demonstrated its value for visual analysis~\cite{wu2018coviar}. Reusing this information can reduce detector workload while maintaining frame-by-frame outputs. The challenge is to preserve localization quality while limiting the cost of preparing motion information and executing updates.

This paper adopts a hardware-software design methodology for edge \acp{GPU}. Its \ac{AI} model for bounding-box estimation is a \ac{CNN} designed around GPU-friendly convolutional operations and independent updates of objects. This structure exposes parallel work across objects and allows the model, \ac{CPU} processing, scheduling, and accelerator assignment to be considered together. The design aims to utilize the available hardware resources efficiently, with its effectiveness evaluated through end-to-end pipeline measurements.

The framework supports two alternative bounding-box estimation methods: analytical motion-vector propagation (Analytical-MV) and CNN-based motion-vector propagation (CNN-MV). Both methods account for object translation and scale variation, including their simultaneous occurrence. Full-frame detector refreshes provide the mechanism for discovering newly visible objects, including those entering from any of the four image borders, whereas propagation updates only previously detected objects.

The main contributions are:
\begin{itemize}
    \item \textbf{Analytical motion model.} A coupled translation-scale estimator propagates the bounding boxes of previously detected objects between full-frame detector refreshes. The refresh mechanism enables the discovery of newly appearing objects.
    \item \textbf{Learned motion model.} An object-local CNN provides an alternative approach for propagating bounding boxes within the same detector-refresh framework.
    \item \textbf{Hardware-aware model and pipeline design.} GPU-friendly convolutions and independent object updates are combined with optimized host processing and evaluated under alternative scheduling and accelerator assignments.
    \item \textbf{Measured operating trade-offs.} Complete-pipeline accuracy, latency, power, energy, and utilization measurements identify the analytical method's efficiency advantage and the conditions favoring learned propagation.
\end{itemize}

\section{Related Work}
\label{sec:related_work}

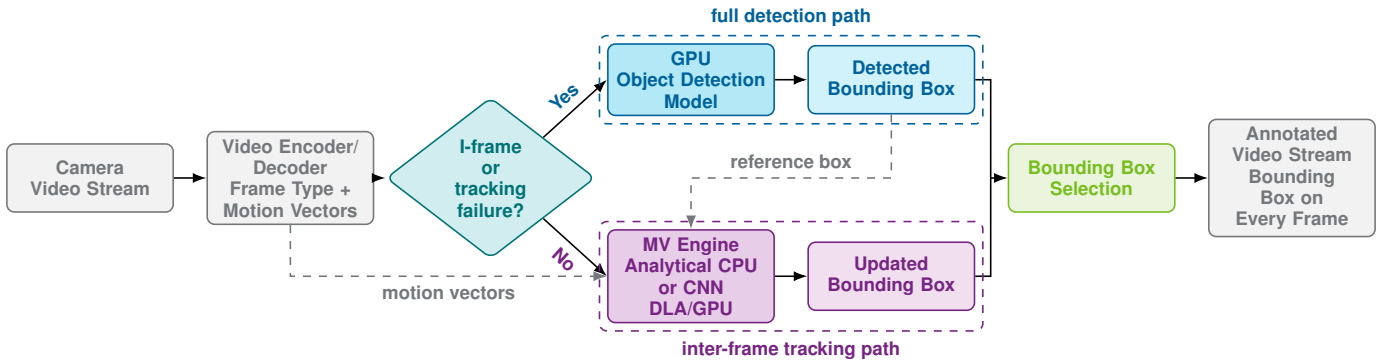
\begin{figure*}[!b]
    \centering
    \setlength{\pipelinecolumnwidth}{\dimexpr(\linewidth-6\diagramgap-1pt)/7\relax}
    \setlength{\pipelinecolumnstep}{\dimexpr\pipelinecolumnwidth+\diagramgap\relax}
    \begin{tikzpicture}[
        x=\pipelinecolumnstep, y=1mm,
        node distance=\diagramgap,
        block/.style={
            diagram block,
            text width=\dimexpr\pipelinecolumnwidth-2mm\relax
        },
        decision/.style={
            diagram block, diamond, aspect=1.3,
            text width=12mm, inner sep=0.5mm,
            minimum height=13mm,
            minimum width=\pipelinecolumnwidth
        },
        arrow/.style={diagram arrow},
        dashedarrow/.style={diagram dashed arrow}
    ]

    \node[block, fill=IEEElgray, draw=IEEEgray, text=IEEEgray]
        (cam) at (0,0) {Camera\\Video Stream};

    \node[block, fill=IEEElgray, draw=IEEEgray, text=IEEEgray]
        (codec) at (1,0) {Video Encoder/\\Decoder\\Frame Type +\\Motion Vectors};

    \node[decision, fill=IEEEteal!18, draw=IEEEdteal, text=IEEEdteal]
        (decide) at (2,0) {I-frame\\or tracking\\failure?};

    \node[block, fill=IEEEcyan!30, draw=IEEEblue, text=IEEEblue]
        (gpu) at (3,13) {GPU\\Object Detection\\Model};

    \node[block, fill=IEEEcyan!18, draw=IEEEblue, text=IEEEblue]
        (detbox) at (4,13) {Detected\\Bounding Box};

    \node[block, fill=IEEEpurple!22, draw=IEEEdpurple, text=IEEEdpurple]
        (mv) at (3,-13) {MV Engine\\Analytical CPU\\or CNN DLA/GPU};

    \node[block, fill=IEEEpurple!14, draw=IEEEdpurple, text=IEEEdpurple]
        (trackbox) at (4,-13) {Updated\\Bounding Box};

    \node[block, fill=IEEEgreen!15, draw=IEEEgreen, text=IEEEgreen]
        (select) at (5,0) {Bounding Box\\Selection};

    \node[block, fill=IEEElgray, draw=IEEEgray, text=IEEEgray]
        (out) at (6,0) {Annotated\\Video Stream\\Bounding Box on\\Every Frame};

    \coordinate (merge) at ($(select.west)+(-2.25mm,0)$);

    \draw[line width=0.6pt, draw=black]
        (detbox.east) -- (merge |- detbox.east) -- (merge);

    \draw[line width=0.6pt, draw=black]
        (trackbox.east) -- (merge |- trackbox.east) -- (merge);

    \draw[arrow] (merge) -- (select.west);

    \draw[arrow] (cam) -- (codec);
    \draw[arrow] (codec) -- (decide);

    \draw[arrow] (decide.north east) -- node[above, sloped,
        font=\diagramfont, text=IEEEblue]
        {Yes} (gpu.west);

    \draw[arrow] (decide.south east) -- node[below, sloped,
        font=\diagramfont, text=IEEEdpurple]
        {No} (mv.west);

    \draw[arrow] (gpu) -- (detbox);
    \draw[arrow] (mv) -- (trackbox);
    \draw[arrow] (select) -- (out);

    \draw[dashedarrow] (codec.south) |- node[pos=0.75, below,
        font=\diagramfont, text=IEEEgray]
        {motion vectors} (mv.west);

    \draw[dashedarrow]
        (detbox.south) -- (detbox.south |- decide.center)
        -- node[above, font=\diagramfont, text=IEEEgray]
        {reference box} (mv.north |- decide.center) -- (mv.north);

    \node[
        diagram group, draw=IEEEblue,
        fit=(gpu)(detbox),
        label={[font=\diagramfont,text=IEEEblue]above:{full detection path}}
    ] {};

    \node[
        diagram group, draw=IEEEdpurple,
        fit=(mv)(trackbox),
        label={[font=\diagramfont,text=IEEEdpurple]below:{inter-frame tracking path}}
    ] {};

    \end{tikzpicture}
    \caption{Motion-vector-assisted video analytics pipeline. The detector path refreshes boxes on I-frames or tracking failures. The motion path uses the CPU Analytical-MV engine or CPU preprocessing followed by CNN-MV inference on DLA/GPU.}
    \label{fig:mv_assisted_video_pipeline}
\end{figure*}

\Ac{RESPIRE}~\cite{tii_dai2022respire} reduces redundant transmission and processing in industrial edge video analytics. Context-aware detection~\cite{tii_zhang2024related} and collaborative edge intelligence~\cite{tii_zhang2023bcei} address complementary recognition and scheduling problems. Our focus is on local box propagation and its complete execution cost.

Majeed \emph{et al.}~\cite{majeed2026scheduling} review heterogeneous edge scheduling, emphasizing memory contention, accelerator compatibility, and transition costs. Sali \emph{et al.}~\cite{sali2025realtime} review real-time object detectors and associated hardware for autonomous vehicles. These surveys motivate joint algorithm-hardware evaluation; our contribution is a measured temporal-propagation pipeline.

\Ac{DFF}~\cite{zhu2017dff} warps key-frame features using optical flow, reporting 73.1\% \ac{mAP} at 20.25~\ac{FPS} on ImageNet \ac{VID}, using an NVIDIA K40 GPU and Intel i7-4790 CPU. \Ac{FGFA}~\cite{zhu2017fgfa} aggregates aligned neighboring features, reaching 76.3\% mAP at 733~ms/frame with a 101-layer residual network (ResNet-101) on a K40 GPU and Intel E5-2670 v2 CPU. Our box-level updates avoid optical-flow inference and feature warping.

Mobile temporal detectors~\cite{liu2018mobile,liu2019looking} show that recurrent inference can be practical on constrained hardware. Looking Fast and Slow~\cite{liu2019looking} interleaves feature extractors with \ac{LSTM} memory; its quantized asynchronous configuration reports 59.3\% mAP at 72.3~FPS on a Pixel 3 phone. CNN-MV instead uses explicit box history and codec vectors without a recurrent hidden state.

\Ac{MVP}~\cite{MVP} uses deterministic motion-vector grids to propagate key-frame detections, reporting 60.9\% mAP at an \ac{IoU} threshold of 0.5 and 10.3~FPS on an NVIDIA RTX 3090 GPU. Its separate translation and scale tests restrict accepted motion. Analytical-MV extends this path with occupied-cell masking and a coupled translationscale fit; the controlled comparison is discussed in Section~\ref{sec:discussion}. See Without Decoding~\cite{seeWithoutDecoding2026} combines box-aligned features with recurrent refinement for compressed-video tracking. Section~\ref{sec:discussion} examines its performance and the distinction from our feed-forward model and embedded measurements.

Because these studies use different datasets and timing protocols, their reported values are not directly comparable with ours. Instead, they motivate the joint treatment of propagation, preprocessing, and resource assignment that follows.

\section{Methodology}
\label{sec:methodology}

\subsection{Pipeline Overview}
\label{subsec:pipeline_overview}

The system alternates full-frame detection with motion-vector propagation (Fig.~\ref{fig:mv_assisted_video_pipeline}). Detection initializes boxes and refreshes them on intra-coded frames (I-frames) or propagation failure, subject to the triggers below. Analytical-MV runs on the CPU; CNN-MV uses CPU preprocessing followed by GPU or \ac{DLA} inference. Both paths validate boxes before emitting each frame's output.

Let $B_t = [x_1,y_1,x_2,y_2]$ denote an object's bounding box in frame $t$, and let $M_{t+1}$ denote the codec motion vectors associated with the transition to frame $t+1$. Analytical-MV estimates box displacement and scale analytically; CNN-MV predicts a candidate box from motion in the object's \ac{ROI} and recent geometry. Both depend on the codec's motion quality and may accumulate errors under occlusion, deformation, or abrupt motion; neither can introduce a previously unseen object without a detector refresh. For CNN-MV, the deployed TensorRT configuration uses the two most recent boxes when history is available:
\begin{equation}
\tilde{B}_{t+1} =
g_\theta(\bar{B}_{t-1},\bar{B}_{t},\bar{M}_{t+1}^{\mathrm{ROI}}),
\end{equation}
where $\bar{B}_{t-1}$ and $\bar{B}_{t}$ are normalized box-history inputs and $\bar{M}_{t+1}^{\mathrm{ROI}}$ is the normalized and padded ROI-filtered motion-vector tensor. The recorded runtime interprets the four output values as a normalized \texttt{xyxy} candidate box, then validates, clips, or rejects it before converting it back to pixel coordinates.

\vspace{-0.5em}
\subsection{Analytical Propagation}
\label{subsec:gridfit_mvp}

The analytical path summarizes backward-reference codec motion vectors over a $3\times3$ grid within each previous-frame box. Raw codec vectors are converted to pixel-domain source and destination centers, and $\mathbf{d}_i$ is defined as the forward displacement from the reference-frame block center to the current-frame block center; for example, a one-pixel rightward block motion gives $d_{x,i}=+1$. The unmodified \ac{MVP} rule is retained as a validation reference, while Analytical-MV excludes empty cells rather than interpreting them as zero motion. Let $\mathcal{O}$ be the set of occupied cells, $\mathbf{g}_i=[g_{x,i},g_{y,i}]^\mathsf{T}$ the center of cell $i$ relative to the bounding-box center, and $\mathbf{d}_i=[d_{x,i},d_{y,i}]^\mathsf{T}$ the mean forward displacement in that cell. Propagation is attempted only when $|\mathcal{O}|\geq2$.

The updated method first tests pure translation. It computes $\bar{\mathbf{d}}=|\mathcal{O}|^{-1}\sum_{i\in\mathcal{O}}\mathbf{d}_i$ and accepts the translation when the standard deviations of both displacement components do not exceed $\tau_{\mathrm{tr}}=1$ pixel. If this test fails, a pure uniform-scale hypothesis about the box center is evaluated using the off-center occupied cells:
\begin{equation}
r_i=\frac{\lVert\mathbf{g}_i+\mathbf{d}_i\rVert_2}{\lVert\mathbf{g}_i\rVert_2}.
\end{equation}
The hypothesis is accepted when $\operatorname{std}(r_i)<\tau_{\mathrm{sc}}=0.1$, with scale $s=\operatorname{mean}(r_i)$ and zero translation.

When neither test explains the motion, we implement a coupled translation and uniform-scale model test,
\begin{equation}
\mathbf{d}_i=\mathbf{t}+\alpha\mathbf{g}_i, \qquad s=1+\alpha,
\end{equation}
where $\mathbf{t}=[t_x,t_y]^\mathsf{T}$. The three unknowns $(t_x,t_y,\alpha)$ are obtained by linear least squares over the occupied cells. A rank-deficient system is rejected. For a full-rank fit, the pixel-domain residual is
\begin{equation}
\varepsilon_{\mathrm{fit}}=
\sqrt{\frac{1}{|\mathcal{O}|}\sum_{i\in\mathcal{O}}
\left\lVert\mathbf{d}_i-\left(\mathbf{t}+\alpha\mathbf{g}_i\right)\right\rVert_2^2},
\end{equation}
and propagation is accepted when $\varepsilon_{\mathrm{fit}}\leq\tau_{\mathrm{fit}}$, with $\tau_{\mathrm{fit}}=10$ pixels in the evaluated implementation. The accepted translation updates the box center, the accepted scale multiplies its width and height, and the resulting normalized box is clamped to the image bounds. If any active box lacks sufficient occupied cells or fails all motion tests, propagation for that frame is rejected, and the object detector is invoked.

\vspace{-0.5em}
\subsection{Object-Local Motion-Vector Selection}
\label{subsec:roi_mv_selection}

CNN-MV preprocessing first filters valid backward-reference codec vectors, fits a robust global similarity transform between source and destination block centers, subtracts that transform to obtain residual motion, and retains coherent residual vectors. The model's residual addition combines the predicted box delta with the previous normalized box; the runtime receives the resulting candidate box, not a delta to add again. Per-object \ac{ROI} selection then reduces each retained record to eight geometric features and pads or truncates the result to the TensorRT binding \texttt{mv=(1,1,2000,8)}. These full-frame and per-object operations account for substantial host work before the compact CNN executes. A representative record is
\begin{equation}
[w,h,x_s,y_s,x_d,y_d,m_x,m_y],
\end{equation}
where $w$ and $h$ are block dimensions, $(x_s,y_s)$ is the source block center, $(x_d,y_d)$ is the destination block center, and $(m_x,m_y)$ is the motion vector or residual motion after background compensation. Non-geometric bookkeeping fields from parser outputs are not supplied to the deployed network. The recorded configuration uses source-coordinate ROI selection: a vector row is retained when its configured source center lies inside the previous box,
\begin{equation}
x_1 \le x_s \le x_2, \qquad y_1 \le y_s \le y_2 .
\end{equation}
Each retained feature is standardized as $(z_j-\mu_j)/\sigma_j$ using training-set statistics stored in the model configuration; only box coordinates are divided by frame dimensions. If fewer than 2000 vectors remain, the tensor is zero-padded; otherwise, it is truncated to the first 2000 rows. The four-value-normalized candidate box is then subjected to the validation gates below.

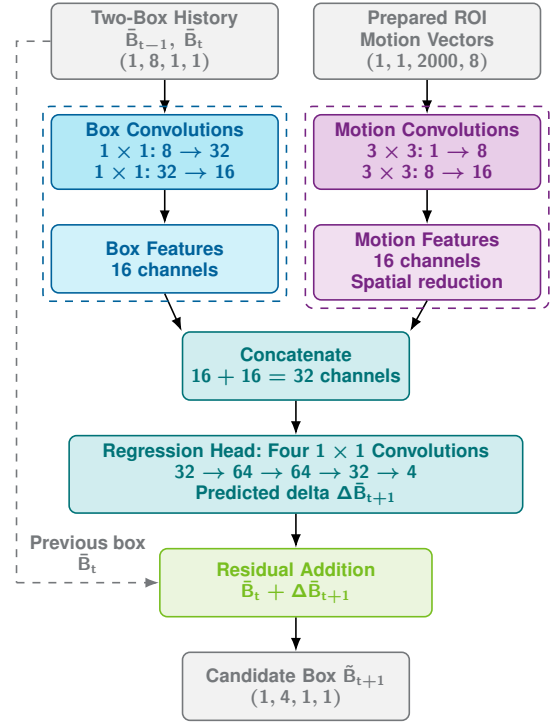
\begin{figure}[t]
    \centering
    \begin{tikzpicture}[
        node distance=\diagramgap,
        block/.style={diagram block, text width=28mm},
        branch/.style={block},
        feature/.style={block},
        arrow/.style={diagram arrow},
        dashedarrow/.style={diagram dashed arrow}
    ]
    \node[block, fill=IEEElgray, draw=IEEEgray, text=IEEEgray]
        (boxes)
        {Two-Box History\\$\mathsf{\bar B_{t-1},\,\bar B_t}$\\$\mathsf{(1,8,1,1)}$};
    \node[branch, below=of boxes, fill=IEEEcyan!30, draw=IEEEblue, text=IEEEblue]
        (boxbranch)
        {Box Convolutions\\$\mathsf{1\times1}$: $\mathsf{8\to32}$\\$\mathsf{1\times1}$: $\mathsf{32\to16}$};
    \node[feature, below=of boxbranch, fill=IEEEcyan!18, draw=IEEEblue, text=IEEEblue]
        (boxfeatures)
        {Box Features\\16 channels};

    \node[block, right=of boxes, fill=IEEElgray, draw=IEEEgray, text=IEEEgray]
        (motion)
        {Prepared ROI\\Motion Vectors\\$\mathsf{(1,1,2000,8)}$};
    \node[branch, below=of motion, fill=IEEEpurple!22, draw=IEEEdpurple, text=IEEEdpurple]
        (motionbranch)
        {Motion Convolutions\\$\mathsf{3\times3}$: $\mathsf{1\to8}$\\$\mathsf{3\times3}$: $\mathsf{8\to16}$};
    \node[feature, below=of motionbranch, fill=IEEEpurple!14, draw=IEEEdpurple, text=IEEEdpurple]
        (motionfeatures)
        {Motion Features\\16 channels\\Spatial reduction};

    \coordinate (branchmerge) at ($(boxfeatures.south)!0.5!(motionfeatures.south)$);
    \node[block, below=of branchmerge, fill=IEEEteal!18, draw=IEEEdteal, text=IEEEdteal]
        (fusion)
        {Concatenate\\$\mathsf{16+16=32}$ channels};

    \node[block, below=of fusion, text width=58mm,
        fill=IEEEteal!18, draw=IEEEdteal, text=IEEEdteal]
        (head)
        {Regression Head: Four $\mathsf{1\times1}$ Convolutions\\$\mathsf{32\to64\to64\to32\to4}$\\Predicted delta $\mathsf{\Delta\bar B_{t+1}}$};
    \node[block, below=of head, text width=34mm,
        fill=IEEEgreen!15, draw=IEEEgreen, text=IEEEgreen]
        (residual)
        {Residual Addition\\$\mathsf{\bar B_t+\Delta\bar B_{t+1}}$};
    \node[block, below=of residual, fill=IEEElgray, draw=IEEEgray, text=IEEEgray]
        (output)
        {Candidate Box $\mathsf{\tilde B_{t+1}}$\\$\mathsf{(1,4,1,1)}$};

    \draw[arrow] (boxes) -- (boxbranch);
    \draw[arrow] (boxbranch) -- (boxfeatures);
    \draw[arrow] (motion) -- (motionbranch);
    \draw[arrow] (motionbranch) -- (motionfeatures);
    \draw[arrow] (boxfeatures.south) -- (fusion.north west);
    \draw[arrow] (motionfeatures.south) -- (fusion.north east);
    \draw[arrow] (fusion) -- (head);
    \draw[arrow] (head) -- (residual);
    \draw[arrow] (residual) -- (output);
    \coordinate (historyroute) at ($(boxes.west)+(-\diagramgap,0)$);
    \draw[dashed, line width=0.6pt, draw=IEEEgray]
        (boxes.west) -- (historyroute) -- (historyroute |- residual.west);
    \draw[dashedarrow] (historyroute |- residual.west) --
        node[above, align=center, font=\diagramfont, text=IEEEgray]
        {Previous box\\$\mathsf{\bar B_t}$} (residual.west);

    \node[diagram group, draw=IEEEblue,
        fit=(boxbranch)(boxfeatures)] {};
    \node[diagram group, draw=IEEEdpurple,
        fit=(motionbranch)(motionfeatures)] {};
    \end{tikzpicture}
    \caption{CNN-MV architecture verified against the model definition and exported graph: 10,548 learned parameters. Dimensions are batch, channels, height, and width. Activation and spatial-reduction details are specified in Section~\ref{subsec:learned_propagation_model}.}
    \label{fig:architecture}
\end{figure}

\vspace{-0.5em}
\subsection{CNN-MV Propagation Model}
\label{subsec:learned_propagation_model}

The DLA- and GPU-targeted TensorRT engines accept box history \texttt{bb=(1,8,1,1)} and ROI motion \texttt{mv=(1,1,2000,8)}, returning the normalized candidate box \texttt{bbox=(1,4,1,1)}. History concatenates \texttt{bbox[t-2]} and \texttt{bbox[t-1]}; the current box fills a missing slot from older history.

Figure~\ref{fig:architecture} summarizes the 10,548-parameter CNN-MV model, verified against its definition and exported \ac{ONNX} graph. Each box-branch convolution is followed by a \ac{ReLU}. Each motion-branch convolution uses unit stride and one-pixel padding, followed by ReLU and $2\times2$ max pooling with stride 2, yielding spatial dimensions $2000\times8\to1000\times4\to500\times2$. Three $5\times1$ average pools with stride $(5,1)$ reduce this to $4\times2$; nearest-neighbor width upsampling produces $4\times4$, followed by $4\times4$ average pooling to $1\times1$. The first three head convolutions use ReLU; the final linear output is added to the latest box: $\tilde B_{t+1}=\bar B_t+\Delta\bar B_{t+1}$.

Both TensorRT engines use \ac{FP16}; the supplied training script uses \ac{FP32} tensors without automatic mixed precision. The exported graph requires 5,769,088 convolutional \acp{MAC} per object, equivalent to 11,538,176 \acp{FLOP} when the multiply and addition counts are counted separately. These counts exclude bias addition, activations, pooling, resizing, residual addition, and host preprocessing.

With fixed padded tensors, propagation scales approximately as $O(N_B)$ for $N_B$ active boxes. Independent object updates permit batching and concurrent scheduling; parallel variants use 16 motion-engine contexts (Table~\ref{tab:configuration_key}). DLA execution requests core 0, with no GPU fallback layers listed in the FP16 engine-creation log. This is build-time partition evidence, not layer-level runtime profiling.

\subsection{Refresh, Validation, and Fallback Logic}
\label{subsec:refresh_validation_fallback}

Propagation updates existing boxes; only a full-frame detector refresh can initialize newly visible objects, including those entering at any image border. Because the evaluated system lacks a dedicated border-search detector, an entering object may remain undetected until the next refresh. The CNN-MV runners invoke the detector on I-frames, on the first detection frame, after five consecutive frames without detections, and when optional periodic refresh is enabled; the recorded runs use the default interval of 0.

On propagation frames, the stationary gate holds a box when residual motion intensity per pixel is below $10^{-4}$; otherwise, CNN-MV proposes a candidate. A candidate is rejected if it is non-finite, empty, less than two pixels wide or high after clipping, less than 65\% visible, outside the area-ratio range $[0.3,3.5]$ relative to the previous box, or shifted by more than the larger of 70 pixels and 1.75 times the previous-box diagonal. Rejected candidates retain the previous box with a score decay of 0.98. Full-frame fallback occurs when the invalid-candidate ratio reaches 0.60 or when the propagated area changes by more than 2.5 relative to the last full-frame detections. The measured tables use this recorded 0.60 policy and exclude results from the later 0.80 policy.

\section{Validation}
\label{sec:experimental_setup}

\begin{table}[!t]
\centering
\caption{Configuration identifiers (IDs). CNN-MV shares one checkpoint and validation policy. Pre/post denotes preprocessing/postprocessing; shading marks the principal operating points.}
\label{tab:configuration_key}
\scriptsize
\renewcommand{\arraystretch}{1.08}
\setlength{\tabcolsep}{2.5pt}
\begin{tabularx}{\columnwidth}{@{}c>{\raggedright\arraybackslash}p{0.27\columnwidth}>{\raggedright\arraybackslash}p{0.18\columnwidth}>{\raggedright\arraybackslash}X@{}}
\toprule
\tablehead
\textbf{ID} & \textbf{Method} & \textbf{Motion execution} & \textbf{Host implementation} \\
\midrule
C0 & YOLO-only & None & Detector on every frame \\
C1 & MVP-YOLO (adapted reference) & CPU analytical & Original MVP grid propagation \\
\keyresult
C2 & Analytical-MV (ours) & CPU analytical & Masked grid plus coupled translationscale fit \\
\familyrule
C3 & CNN-MV (ours) & DLA & Serial original \\
C4 & CNN-MV (ours) & DLA & Serial optimized pre/post \\
C5 & CNN-MV (ours) & DLA & Parallel original \\
C6 & CNN-MV (ours) & DLA & Parallel optimized pre/post \\
\familyrule
C7 & CNN-MV (ours) & GPU & Serial original \\
\keyresult
C8 & CNN-MV (ours) & GPU & Serial optimized pre/post \\
C9 & CNN-MV (ours) & GPU & Parallel original \\
C10 & CNN-MV (ours) & GPU & Parallel optimized pre/post \\
\bottomrule
\end{tabularx}
\end{table}

The \ac{UA-DETRAC} traffic-video benchmark~\cite{wen2020uadetrac} is used to evaluate the framework in a controlled traffic-monitoring setting. Its fixed-camera road scenes represent an edge-based intelligent transportation application in which continuous vehicle localization must be performed within limited compute and power budgets. The same propagation and heterogeneous-execution principles may also apply to autonomous vehicles, inspection drones, and mobile robots; however, validation under moving-camera conditions is outside the scope of the present study. Version 11 of \ac{YOLO} (YOLOv11) is the experimental key-frame detector and is not part of the propagation architecture. Other detectors that return conventional boxes could be substituted, but would require separate accuracy, refresh policy, TensorRT, and timing validation.

MVP-YOLO adapts the released MVP implementation~\cite{MVP} using the common YOLO TensorRT engine while retaining its original, untrained grid-propagation rule. It is distinct from our Analytical-MV and CNN-MV methods.

\vspace{-0.5em}
\subsection{Videos and Baselines}
\label{subsec:videos_and_baselines}

Table~\ref{tab:configuration_key} defines the 11 configurations evaluated on the same 25-video subset. A YOLOv11 detector trained on UA-DETRAC serves as the common full-frame detector. All configurations share the same videos, annotations, class list, trained YOLOv11 TensorRT engine, a confidence threshold of 0.5, an \ac{NMS} IoU of 0.45, a maximum of 300 detections, and a 20-frame warm-up. Consequently, differences among the configurations arise from choices in propagation, routing, and execution rather than from the detector. Each benchmark processes 28{,}177 frames; removing 20 warm-up frames from each video leaves 27{,}677 timed rows. Accuracy is evaluated on 27{,}893 annotated frames.

\vspace{-0.5em}
\subsection{Training}
\label{subsec:training}

CNN-MV training samples are constructed from six UA-DETRAC sequences that are distinct from all 25 sequences used for the final evaluation. Samples contain adjacent annotated boxes and motion-vector files, with frame-normalized box coordinates, pairwise IoU of at least 0.25, and matching classes. Within each of the six training sequences, samples are randomly divided using seed 7 and a validation fraction of 0.2, yielding 40{,}167 training and 10{,}042 validation samples. Although this internal validation split is sample-level and may contain temporally adjacent samples across its partitions, the final 25-video system evaluation is sequence-disjoint from the CNN-MV training data.

Training uses the Adam optimizer for 100 epochs, a batch size of 64, a learning rate of 0.001, zero weight decay, Smooth L1 loss with beta 1.0, 2000 motion-vector rows, source-coordinate ROI selection, residual-vector postprocessing, and stationary-box gating at $10^{-4}$. The best checkpoint occurs at epoch 94, with validation loss $1.3049\times10^{-6}$. The training script, checkpoint configuration, exported graph, and normalization statistics document the model configuration; a reproducible release should also include the sample-split manifests.

\vspace{-0.5em}
\subsection{Reference Protocols}
\label{subsec:reference_protocols}

Predictions are evaluated against UA-DETRAC annotations using class-agnostic one-to-one vehicle matching at IoU $\geq 0.5$, processed in descending IoU order. The logged \texttt{car}, \texttt{van}, \texttt{bus}, and \texttt{others} categories are collapsed before matching, and predictions with at least 50\% of their area inside an official ignore region are excluded. Metrics comprise precision, recall, the \ac{F1} score, mean matched IoU, and counts of \acp{TP}, \acp{FP}, and \acp{FN}.

Routing counts use all 28{,}177 processed frames; stationary holds count as MV-routed. YOLO-only invokes the detector on every frame.

Standard \ac{AP} at IoU 0.50 and 0.75 (AP$_{50}$, AP$_{75}$), mAP over 0.500.95 (mAP$_{50:95}$), and per-class AP require confidence sweeps. Retained scores are truncated at 0.5, and non-CNN prediction archives are incomplete, precluding those metrics. Reruns must preserve routing, retain low-score predictions, and use class-aware, confidence-ranked matching.

\subsection{On-Device Measurement Setup}
\label{subsec:ondevice_setup}

On-device experiments use an NVIDIA Jetson AGX Orin Developer Kit with TensorRT 10.3. The live reader decodes the stored UA-DETRAC videos and extracts codec vectors, so decoding remains within the measured pipeline; the experiments do not demonstrate a decoder bypass. Under MAX\_FREQ setting, the recorded CPU, GPU, \ac{EMC}, and DLA maxima are 2.2016~GHz, 1.3005~GHz, 3.199~GHz, and 1.6~GHz, respectively.

The common YOLO detector and all CNN-MV engines use FP16 inference. YOLO exposes \texttt{images=(1,3,960,960)} and \texttt{output0=(1,8,18900)}. DLA variants request core 0, while GPU variants use a separately built plan for the same model. The DLA engine-creation log reports no GPU fallback layers~\cite{nvidiaTensorRTDLA}; YOLO remains on the GPU in every configuration.

Improved host processing batches active boxes, vectorizes residual-neighbor construction, batches box validation and history matching, reuses geometry, and removes intermediate work without changing model inputs or routing. Parallel improved variants add multithreaded per-object scheduling. \Ac{RANSAC}/\ac{IRLS} preprocessing uses a 2.5-pixel fit threshold, at least 20 inliers, a 1.5-pixel residual threshold, a 2-pixel residual-coherence tolerance, support of at least 3 vectors (including the candidate) in its grid cell and neighboring cells, and an empty fallback when no coherent residual support remains.

Each method runs after a reboot. The measured \texttt{-no-video} pass begins timing at local frame read/decode and ends after detection postprocessing. Tegrastats~\cite{nvidiaTegrastats} is active only from the first post-warm-up frame through final measured-frame postprocessing. Model loading, warm-up, annotation drawing, \ac{MPEG}-4 Part 14 (MP4) video-file generation, and result serialization are excluded; annotated videos are generated in a separate unmonitored pass.

Mean power sums complete samples from \texttt{VDD\_GPU\_SOC} (GPU and \ac{SoC}), \texttt{VDD\_CPU\_CV} (CPU and \ac{CV} engines), and \texttt{VIN\_SYS\_5V0} (5-volt system input) within the measured window. Energy per frame is mean rail-sum power divided by effective throughput, without idle subtraction, and is not wall-socket energy. For stage estimates, each frame is assigned the nearest telemetry sample. Stage energy is the mean of frame-matched power multiplied by stage duration; stage power is the corresponding duration-weighted mean. These allocations are not independent electrical measurements of individual stages. YOLO and motion inference are subcomponents of inference/compute and must not be added to it again.

\section{Results}
\label{sec:results}

All results use the configuration IDs in Table~\ref{tab:configuration_key}. Tables separate the non-CNN configurations (C0-C2), CNN-MV/DLA (C3-C6), and CNN-MV/GPU (C7-C10); shaded rows identify Analytical-MV (C2) and serial optimized CNN-MV/GPU (C8). Latency is summarized by the mean, 95th percentile (P95), and 99th percentile (P99); FPS denotes throughput, and FPS/W denotes throughput per watt.

\vspace{-0.5em}
\subsection{Accuracy and Routing}

\begin{table*}[!t]
\centering
\caption{Accuracy and routing for the 11 configurations. Every configuration uses 25 videos, 27,893 annotated frames, and 28,177 processed frames. Prec. denotes precision; IoU is averaged over matched pairs. Detector and MV percentages use the processed-frame total.}
\label{tab:accuracy_routing_results}
\footnotesize
\renewcommand{\arraystretch}{1.12}
\setlength{\tabcolsep}{3.1pt}
\begin{tabular*}{\textwidth}{@{\extracolsep{\fill}}lrrrrrrrrrrr@{}}
\toprule
\tablehead
\textbf{ID} & \multicolumn{4}{c}{\textbf{Accuracy}} & \multicolumn{3}{c}{\textbf{Detection counts}} & \multicolumn{2}{c}{\textbf{Detector routing}} & \multicolumn{2}{c}{\textbf{MV routing}} \\
\cmidrule(lr){2-5}\cmidrule(lr){6-8}\cmidrule(lr){9-10}\cmidrule(lr){11-12}
 & Prec. & Recall & F1 & IoU & TP & FP & FN & Frames & \% & Frames & \% \\
\midrule
C0  & \textbf{0.957} & \textbf{0.910} & \textbf{0.933} & \textbf{0.904} & \textbf{197,005} & 8,940 & \textbf{19,522} & 28,177 & 100.0 & 0 & 0.0 \\
C1  & \textbf{0.957} & 0.874 & 0.914 & 0.901 & 189,343 & \textbf{8,594} & 27,184 & 21,048 & 74.7 & 7,129 & 25.3 \\
\keyresult
C2  & 0.951 & 0.871 & 0.909 & 0.880 & 188,543 & 9,708 & 27,984 & 13,515 & 48.0 & \textbf{14,662} & \textbf{52.0} \\
\familyrule
C3  & 0.934 & 0.890 & 0.911 & 0.874 & 192,795 & 13,730 & 23,732 & 15,647 & 55.5 & 12,530 & 44.5 \\
C4  & 0.934 & 0.890 & 0.911 & 0.874 & 192,795 & 13,730 & 23,732 & 15,647 & 55.5 & 12,530 & 44.5 \\
C5  & 0.933 & 0.890 & 0.911 & 0.874 & 192,814 & 13,796 & 23,713 & 15,739 & 55.9 & 12,438 & 44.1 \\
C6  & 0.934 & 0.890 & 0.911 & 0.874 & 192,795 & 13,730 & 23,732 & 15,646 & 55.5 & 12,531 & 44.5 \\
\familyrule
C7  & 0.934 & 0.890 & 0.911 & 0.874 & 192,793 & 13,732 & 23,734 & 15,646 & 55.5 & 12,531 & 44.5 \\
\keyresult
C8  & 0.934 & 0.890 & 0.911 & 0.874 & 192,794 & 13,731 & 23,733 & 15,646 & 55.5 & 12,531 & 44.5 \\
C9  & 0.933 & 0.891 & 0.911 & 0.874 & 192,835 & 13,775 & 23,692 & 15,739 & 55.9 & 12,438 & 44.1 \\
C10 & 0.934 & 0.890 & 0.911 & 0.874 & 192,793 & 13,732 & 23,734 & 15,646 & 55.5 & 12,531 & 44.5 \\
\bottomrule
\end{tabular*}
\end{table*}

YOLO-only leads in F1 (0.933), followed by MVP-YOLO (0.914). Analytical-MV reaches 0.909, trading approximately 2.4 percentage points for efficiency. CNN-MV variants cluster around precision/recall/F1 of 0.934/0.890/0.911. Relative to Analytical-MV, they gain approximately 1.9 percentage points of recall and 0.2 percentage points of F1, with lower precision and matched IoU (Table~\ref{tab:accuracy_routing_results}).

MVP-YOLO propagates 25.3\% of frames, Analytical-MV 52.0\%, and CNN-MV 44.1-44.5\%, including stationary holds under the 0.60 fallback policy. Equivalently, Analytical-MV reduces the fraction of detector-processed frames from 100\% for framewise YOLO and 74.7\% for MVP-YOLO to 48.0\%, while CNN-MV reduces it to approximately 55.5\%. The different routing rates are intentional outcomes of the complete propagation and fallback policies. These comparisons, therefore, characterize complete-system operating points rather than propagation accuracy under a matched detector budget.

\vspace{-0.5em}
\subsection{Latency and Per-Video Variation}

\begin{table*}[!t]
\centering
\caption{Complete-pipeline timing for the 11 configurations. Latencies are milliseconds; throughput excludes 20 warm-up frames per video. Engine is the mean per-call engine latency, Compute is synchronized per-frame inference time, and Calls/frame is the average number of detector or motion-engine invocations. For CNN-MV, Engine FPS is based on Compute because multiple motion-engine calls may occur per frame.}
\label{tab:latency_results}
\footnotesize
\renewcommand{\arraystretch}{1.12}
\setlength{\tabcolsep}{4pt}
\begin{tabular*}{\textwidth}{@{\extracolsep{\fill}}lrrrrrrrr@{}}
\toprule
\tablehead
\textbf{ID} & \multicolumn{2}{c}{\textbf{Throughput (FPS)}} & \multicolumn{3}{c}{\textbf{End-to-end latency (ms)}} & \multicolumn{2}{c}{\textbf{Inference time (ms)}} & \textbf{Calls/} \\
\cmidrule(lr){2-3}\cmidrule(lr){4-6}\cmidrule(lr){7-8}
 & Pipeline & Engine & Mean & P95 & P99 & Engine/call & Compute/frame & \textbf{frame} \\
\midrule
C0  & 82.00 & 249.31 & 12.20 & \textbf{13.28} & \textbf{13.68} & 4.01 & 4.01 & 1.000 \\
C1  & 85.33 & 247.99 & 11.72 & 15.90 & 18.39 & 4.03 & 4.25 & 0.747 \\
\keyresult
C2  & \textbf{110.69} & 249.16 & \textbf{9.03} & 16.90 & 18.24 & 4.01 & 3.44 & \textbf{0.479} \\
\familyrule
C3  & 11.67 & 193.10 & 85.71 & 124.63 & 138.71 & 0.37 & 5.18 & 6.017 \\
C4  & 48.46 & 194.94 & 20.64 & 36.20 & 41.55 & 0.37 & 5.13 & 6.017 \\
C5  & 11.70 & 171.53 & 85.44 & 121.97 & 134.85 & \textbf{0.36} & 5.83 & 6.213 \\
C6  & 51.96 & 178.94 & 19.25 & 32.06 & 36.44 & 0.37 & 5.59 & 6.017 \\
\familyrule
C7  & 12.05 & \textbf{339.11} & 82.99 & 118.24 & 129.95 & 0.37 & \textbf{2.95} & 6.017 \\
\keyresult
C8  & 54.29 & 339.08 & 18.42 & 30.30 & 33.59 & 0.37 & \textbf{2.95} & 6.017 \\
C9  & 11.84 & 191.70 & 84.46 & 120.72 & 132.90 & \textbf{0.36} & 5.22 & 6.214 \\
C10 & 53.08 & 198.09 & 18.84 & 31.52 & 35.45 & 0.37 & 5.05 & 6.017 \\
\bottomrule
\end{tabular*}
\end{table*}

Analytical-MV leads at 110.69~FPS and 9.03~ms mean latency: throughput improves by 35.0\%, and latency falls by 25.9\% relative to YOLO-only. Its P95 is nevertheless higher (16.90 versus 13.28~ms), reflecting expensive detector-recovery frames (Table~\ref{tab:latency_results}).

Serial-improved GPU execution is the fastest CNN-MV variant (54.29 FPS, 18.42 ms mean, 30.30 ms P95). Against serial-improved DLA execution, it reduces mean latency by 10.7\%, P95 by 16.3\%, and energy by 11.4\%, with an F1 difference below 0.00001. Parallel scheduling improves optimized DLA latency by 6.7\% to 19.25~ms but slightly slows optimized GPU execution. Original CNN-MV variants remain host-bound at 11.67-12.05~FPS.

Table~\ref{tab:route_latency_results} separates propagation from detector recovery/fallback: mean costs are 3.68 versus 14.99~ms for Analytical-MV and 11.63 versus 24.32~ms for serial-improved GPU CNN-MV.

\begin{table*}[!t]
\centering
\caption{Route-specific timing in milliseconds. Motion is the per-frame motion-compute contribution; propagation and recovery/fallback columns report complete-frame latency conditioned on the route. YOLO is mean detector-engine latency per call. Calls/frame includes detector and motion-engine invocations; C0 has no propagation route.}
\label{tab:route_latency_results}
\footnotesize
\renewcommand{\arraystretch}{1.12}
\setlength{\tabcolsep}{4pt}
\begin{tabular*}{\textwidth}{@{\extracolsep{\fill}}lrrrrrrrr@{}}
\toprule
\tablehead
\textbf{ID} & \textbf{YOLO} & \multicolumn{2}{c}{\textbf{Motion compute}} & \multicolumn{2}{c}{\textbf{Propagation route}} & \multicolumn{2}{c}{\textbf{YOLO recovery/fallback}} & \textbf{Calls/} \\
\cmidrule(lr){3-4}\cmidrule(lr){5-6}\cmidrule(lr){7-8}
 & Mean & Mean & P95 & Mean & P95 & Mean & P95 & \textbf{frame} \\
\midrule
C1  & 4.03 & 1.24 & 3.74 & 3.93 & 8.25 & 14.43 & 16.34 & 0.747 \\
\keyresult
C2  & 4.01 & 1.52 & 3.32 & 3.68 & 6.65 & 14.99 & 17.66 & 0.479 \\
\familyrule
C3  & 4.00 & 2.96 & 8.05 & 76.83 & 92.85 & 94.19 & 130.32 & 6.017 \\
C4  & 4.01 & 2.91 & 7.97 & 13.40 & 19.79 & 27.07 & 38.30 & 6.017 \\
C5  & 4.00 & 3.60 & 8.69 & 76.80 & 91.31 & 93.34 & 127.11 & 6.213 \\
C6  & 4.02 & 3.36 & 8.18 & 12.36 & 16.79 & 25.29 & 33.89 & 6.017 \\
\familyrule
C7  & 4.01 & 0.72 & 2.03 & 74.66 & 87.94 & 90.83 & 122.94 & 6.017 \\
\keyresult
C8  & 3.99 & 0.74 & 2.03 & 11.63 & 15.49 & 24.32 & 31.64 & 6.017 \\
C9  & 4.02 & 2.97 & 7.62 & 75.88 & 89.84 & 92.22 & 125.54 & 6.214 \\
C10 & 4.00 & 2.83 & 7.34 & 11.87 & 16.20 & 24.90 & 33.23 & 6.017 \\
\bottomrule
\end{tabular*}
\end{table*}

\begin{figure*}[!t]
    \centering
    \includegraphics[width=\textwidth]{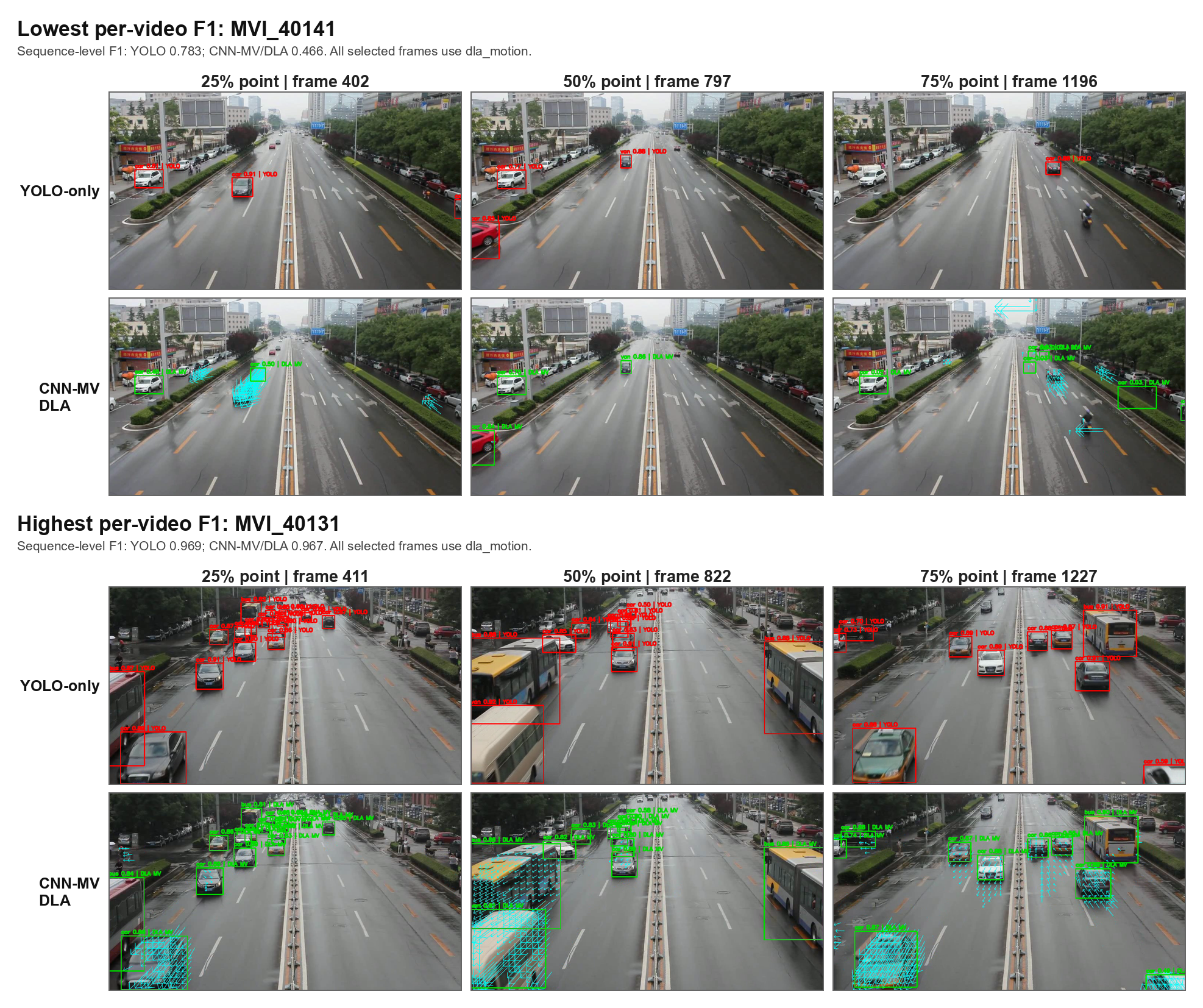}
    \caption{YOLO-only versus serial-improved CNN-MV/DLA on successful motion frames nearest each sequence's quartiles: MVI\_40141 frames 402/797/1196 (sequence F1: 0.783/0.466) and MVI\_40131 frames 411/822/1227 (0.969/0.967). F1 pairs are YOLO/CNN-MV. Outputs lack ground-truth overlays and do not establish per-frame accuracy.}
    \label{fig:mv_dla_accuracy_examples}
\end{figure*}

For the serial-improved GPU CNN-MV, per-video throughput ranges from 38.73-93.58~FPS, F1 ranges from 0.466-0.967, and energy ranges from 0.166-0.453~J/frame. The optimized serial DLA variant shares the lowest- and highest-F1 sequences, MVI\_40141 and MVI\_40131. Figure~\ref{fig:mv_dla_accuracy_examples} compares their outputs with YOLO-only, illustrating the variation hidden by aggregate metrics.

\subsection{Power, Energy, and Utilization}

Under the rail-sum measurement protocol, optimized CNN-MV draws 17.20-17.79~W, below all non-CNN configurations. Original CNN-MV draws only 13.58-14.14~W, but prolonged host processing raises energy to 1.127-1.212~J/frame (Table~\ref{tab:power_results}).

Serial-improved GPU CNN-MV leads its family at 0.319~J/frame and 3.134~FPS/W; Analytical-MV remains better at 0.177~J/frame and 5.637~FPS/W. Serial-improved DLA execution lowers mean GPU utilization from 15.5\% to 11.8\% (Table~\ref{tab:utilization_results}), a proxy for headroom that requires a concurrent-workload experiment to assess its sharing benefit.

\begin{table*}[!t]
\centering
\caption{Measured rail power and estimated energy efficiency. Total power is the three-rail sum. Frame mean assigns the nearest telemetry sample to each timed frame; energy uses the measured-window mean divided by pipeline FPS, with no idle subtraction.}
\label{tab:power_results}
\footnotesize
\renewcommand{\arraystretch}{1.12}
\setlength{\tabcolsep}{3.8pt}
\begin{tabular*}{\textwidth}{@{\extracolsep{\fill}}lrrrrrrrr@{}}
\toprule
\tablehead
\textbf{ID} & \multicolumn{2}{c}{\textbf{Total power (W)}} & \multicolumn{3}{c}{\textbf{Mean rail power (W)}} & \textbf{Frame mean} & \multicolumn{2}{c}{\textbf{Energy efficiency}} \\
\cmidrule(lr){2-3}\cmidrule(lr){4-6}\cmidrule(lr){8-9}
 & Mean & P95 & GPU-SoC & CPU-CV & 5V & (W) & J/frame & FPS/W \\
\midrule
C0  & 22.88 & 23.54 & 12.98 & 3.19 & 6.70 & 22.90 & 0.279 & 3.584 \\
C1  & 20.60 & 21.73 & 11.28 & 3.00 & 6.33 & 20.50 & 0.241 & 4.142 \\
\keyresult
C2  & 19.64 & 21.24 & 10.30 & 3.15 & 6.19 & 19.50 & \textbf{0.177} & \textbf{5.637} \\
\familyrule
C3  & 14.14 & 14.73 & 5.92 & 3.19 & 5.04 & 14.23 & 1.212 & 0.825 \\
C4  & 17.45 & 18.63 & 8.18 & 3.45 & 5.82 & 17.33 & 0.360 & 2.777 \\
C5  & 14.12 & 14.73 & 5.90 & 3.19 & 5.03 & 14.22 & 1.206 & 0.829 \\
C6  & 17.79 & 18.94 & 8.33 & 3.56 & 5.90 & 17.65 & 0.342 & 2.920 \\
\familyrule
C7  & \textbf{13.58} & \textbf{14.13} & \textbf{5.90} & \textbf{2.79} & \textbf{4.89} & \textbf{13.68} & 1.127 & 0.887 \\
\keyresult
C8  & 17.32 & 18.62 & 8.48 & 3.17 & 5.67 & 17.21 & 0.319 & 3.134 \\
C9  & 13.62 & \textbf{14.13} & \textbf{5.90} & \textbf{2.79} & 4.92 & 13.72 & 1.150 & 0.870 \\
C10 & 17.20 & 18.52 & 8.43 & 3.15 & 5.62 & 17.09 & 0.324 & 3.086 \\
\bottomrule
\end{tabular*}
\end{table*}

\begin{table*}[!t]
\centering
\caption{Utilization and temperature over the measured telemetry windows. Utilization is the time-averaged recorded percentage; temperatures are in degrees Celsius. A dash denotes unavailable telemetry, while 0.0 is a recorded, rounded value and does not establish the absence of short accelerator activity.}
\label{tab:utilization_results}
\footnotesize
\renewcommand{\arraystretch}{1.12}
\setlength{\tabcolsep}{4pt}
\begin{tabular*}{\textwidth}{@{\extracolsep{\fill}}lrrrrrrrr@{}}
\toprule
\tablehead
\textbf{ID} & \multicolumn{4}{c}{\textbf{Mean utilization (\%)}} & \multicolumn{2}{c}{\textbf{CPU temperature ($^\circ$C)}} & \multicolumn{2}{c}{\textbf{GPU temperature ($^\circ$C)}} \\
\cmidrule(lr){2-5}\cmidrule(lr){6-7}\cmidrule(lr){8-9}
 & CPU & GPU & DLA0 & EMC & Mean & Max. & Mean & Max. \\
\midrule
C0  & 9.2 & 33.1 & - & 13.9 & 56.3 & 59.0 & 52.8 & 55.0 \\
C1  & 8.9 & 25.8 & - & 10.8 & 57.6 & 58.8 & 54.0 & 55.3 \\
\keyresult
C2  & 9.7 & 21.8 & - & 9.2 & 57.1 & 58.8 & 52.9 & 54.2 \\
\familyrule
C3  & 8.1 & 2.9 & 0.0 & 1.3 & 54.9 & 56.4 & 50.4 & 52.1 \\
C4  & 12.1 & 11.8 & 1.4 & 6.1 & 56.8 & 58.3 & 52.2 & 53.7 \\
C5  & 8.3 & 3.1 & 0.0 & 1.4 & 54.7 & 56.9 & 50.3 & 52.0 \\
C6  & 12.9 & 13.0 & 2.0 & 6.6 & 56.8 & 58.4 & 52.2 & 53.5 \\
\familyrule
C7  & 8.3 & 3.6 & 0.0 & 1.0 & 54.3 & 56.1 & 49.8 & 51.5 \\
\keyresult
C8  & 13.0 & 15.5 & 0.0 & 5.4 & 56.2 & 58.3 & 51.9 & 53.3 \\
C9  & 8.3 & 3.3 & 0.0 & 1.0 & 54.2 & 55.5 & 49.9 & 51.6 \\
C10 & 12.9 & 14.9 & 0.0 & 5.3 & 55.7 & 58.3 & 51.5 & 53.2 \\
\bottomrule
\end{tabular*}
\end{table*}

\subsection{Stage-Level Analysis}

Host preprocessing dominates CNN-MV (Table~\ref{tab:stage_results}). Optimization reduces it from 73.54 to 9.09~ms on DLA and from 73.50 to 9.08~ms on GPU, lowering end-to-end latency by 75.9\% and 77.8\%, respectively, and energy by 70.3\% and 71.7\%, without materially changing accuracy.

A preliminary \texttt{CuPy} preprocessing path increased latency through small-tensor transfers and was discarded. CPU batching/vectorization was retained, with multithreading in parallel variants. GPU placement reduces optimized serial compute time from 5.13 to 2.95~ms/frame.

\begin{figure*}[!t]
    \centering
    \includegraphics[width=\textwidth]{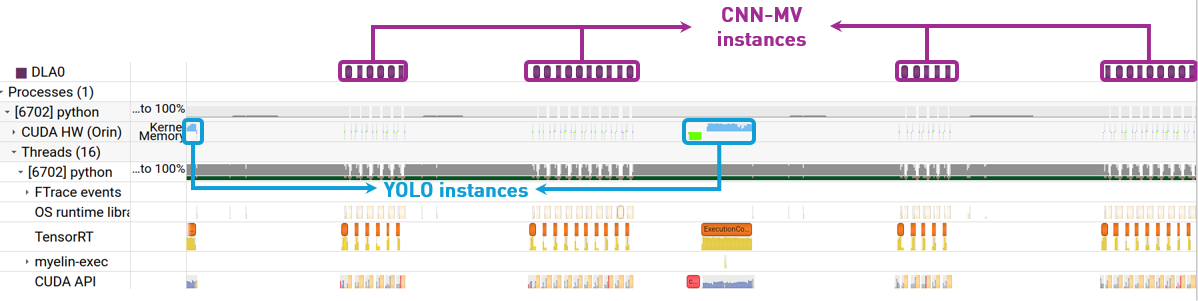}
    \caption{Nsight Systems timeline: CNN-MV on DLA core 0 (purple) and YOLO on GPU (blue), with thread, TensorRT, \ac{CUDA} \ac{API}, and \ac{OS} activity below.}
    \label{fig:nsight_systems_timeline}
\end{figure*}

Figure~\ref{fig:nsight_systems_timeline} illustrates the temporal separation of CNN-MV activity on DLA core 0 and YOLO activity on the GPU.

\begin{table*}[!t]
\centering
\caption{Mean component latency in milliseconds per frame. Compute includes YOLO and motion-model inference, which are shown separately for interpretation and must not be added again. Overhead contains synchronization, scheduling, and residual bookkeeping; in parallel variants, it is reported after any overlap already present in the measured frame time.}
\label{tab:stage_results}
\footnotesize
\renewcommand{\arraystretch}{1.12}
\setlength{\tabcolsep}{4pt}
\begin{tabular*}{\textwidth}{@{\extracolsep{\fill}}lrrrrrrr@{}}
\toprule
\tablehead
\textbf{ID} & \textbf{Decode} & \textbf{Preprocess} & \multicolumn{3}{c}{\textbf{Compute (total and components)}} & \textbf{Postprocess} & \textbf{Overhead} \\
\cmidrule(lr){4-6}
 & & & Total & YOLO & Motion & & \\
\midrule
C0  & 0.60 & 3.90 & 4.01 & 4.01 & - & 1.69 & 2.00 \\
C1  & 1.64 & 2.84 & 4.25 & 3.01 & 1.24 & 1.32 & 1.67 \\
\keyresult
C2  & 1.61 & \textbf{1.83} & 3.44 & 1.92 & 1.52 & \textbf{0.87} & 1.27 \\
\familyrule
C3  & 1.67 & 73.54 & 5.18 & 2.22 & 2.96 & 2.29 & 3.03 \\
C4  & 1.65 & 9.09 & 5.13 & 2.22 & 2.91 & 1.80 & 2.97 \\
C5  & 1.68 & 74.68 & 5.83 & 2.23 & 3.60 & 2.20 & \textbf{1.06} \\
C6  & 1.67 & 9.10 & 5.59 & 2.23 & 3.36 & 1.83 & \textbf{1.06} \\
\familyrule
C7  & 1.68 & 73.50 & \textbf{2.95} & 2.23 & \textbf{0.72} & 1.91 & 2.96 \\
\keyresult
C8  & 1.65 & 9.08 & \textbf{2.95} & \textbf{2.21} & 0.74 & 1.78 & 2.96 \\
C9  & 1.67 & 74.53 & 5.22 & 2.24 & 2.97 & 1.91 & 1.13 \\
C10 & 1.66 & 9.18 & 5.05 & 2.22 & 2.83 & 1.84 & 1.12 \\
\bottomrule
\end{tabular*}
\end{table*}

Table~\ref{tab:stage_energy_results} allocates frame-matched rail power according to each stage's measured duration. Optimized serial preprocessing energy falls from 1038.7 to 158.1~mJ/frame on DLA and from 997.3 to 156.8~mJ/frame on GPU. The corresponding aggregate and rail-level power measurements remain available in Table~\ref{tab:power_results}; small differences between stage totals and aggregate energy are due to frame matching and stage accounting.

\begin{table*}[!t]
\centering
\caption{Estimated stage energy (mJ/frame), averaged from frame-matched rail power multiplied by recorded stage duration. YOLO and Motion are included in Compute and do not need to be added again. A dash indicates no motion stage. These are telemetry-based allocations without idle subtraction; they are not direct component measurements.}
\label{tab:stage_energy_results}
\footnotesize
\renewcommand{\arraystretch}{1.12}
\setlength{\tabcolsep}{4pt}
\begin{tabular*}{\textwidth}{@{\extracolsep{\fill}}lrrrrrrr@{}}
\toprule
\tablehead
\textbf{ID} & \textbf{Decode} & \textbf{Preprocess} & \multicolumn{3}{c}{\textbf{Compute (total and components)}} & \textbf{Postprocess} & \textbf{Overhead} \\
\cmidrule(lr){4-6}
 & & & Total & YOLO & Motion & & \\
\midrule
C0  & 13.6 & 89.3 & 91.8 & 91.8 & - & 38.4 & 45.9 \\
C1  & 33.7 & 58.6 & 87.5 & 62.3 & 25.2 & 27.0 & 34.4 \\
\keyresult
C2  & 31.5 & 36.4 & 67.5 & 38.2 & 29.4 & 17.1 & 25.0 \\
\familyrule
C3  & 23.8 & 1038.7 & 74.1 & 32.2 & 41.9 & 32.6 & 43.2 \\
C4  & 28.6 & 158.1 & 90.0 & 39.5 & 50.5 & 31.3 & 51.9 \\
C5  & 23.9 & 1053.3 & 83.2 & 32.3 & 50.8 & 31.2 & 15.3 \\
C6  & 29.4 & 161.3 & 100.1 & 40.3 & 59.8 & 32.6 & 19.2 \\
\familyrule
C7  & 22.9 & 997.3 & 40.9 & 31.1 & 9.8 & 26.1 & 40.6 \\
\keyresult
C8  & 28.3 & 156.8 & 51.8 & 39.1 & 12.7 & 30.9 & 51.5 \\
C9  & 22.9 & 1013.3 & 71.9 & 31.4 & 40.5 & 26.2 & 15.8 \\
C10 & 28.4 & 157.3 & 87.3 & 38.9 & 48.5 & 31.7 & 19.5 \\
\bottomrule
\end{tabular*}
\end{table*}

\section{Discussion and Comparison with Prior Work}
\label{sec:discussion}

Our framework targets continuous vehicle localization for edge-based traffic monitoring and operates on detector outputs within one edge platform. This differs from RESPIRE's transmission/processing reduction~\cite{tii_dai2022respire}, contextual recognition inside a detector~\cite{tii_zhang2024related}, and cross-device collaborative scheduling~\cite{tii_zhang2023bcei}.

The scheduling review of Majeed \emph{et al.}~\cite{majeed2026scheduling} explains why accelerator assignment must account for contention and transition overhead. Our measurements complement that perspective by separating host optimization from whole-model DLA/GPU placement; we do not propose a layer-partitioning scheduler. The detector-hardware survey by Sali \emph{et al.}~\cite{sali2025realtime} addresses autonomous-vehicle perception more broadly, whereas our evaluation focuses on temporal box reuse on a single edge platform.

Like DFF and FGFA, \ac{LSTS}~\cite{jiang2019lsts}, \ac{PSLA}~\cite{guo2019psla}, \ac{SELSA}~\cite{wu2019selsa}, and \ac{MEGA}~\cite{chen2020mega} retain richer visual features through temporal alignment or aggregation. Our box updates avoid this intermediate computation but require detector refresh for discovery and recovery. Different datasets and mAP/F1 protocols preclude a direct ranking of accuracy.

Against MVP~\cite{MVP}, Analytical-MV adds occupied-cell masking and coupled translation-scale fitting. With a common detector, it reduces MVP-YOLO's latency from 11.72 to 9.03~ms and energy from 0.241 to 0.177~J/frame, with F1 changing from 0.914 to 0.909. CNN-MV adds a learned, nonrecurrent alternative and measured host/accelerator trade-offs.

See Without Decoding~\cite{seeWithoutDecoding2026} combines \ac{BAFE} with \ac{BiLSTM} refinement, reporting 89.62\% mAP at IoU 0.5 on static-camera Multi-Object Tracking 2017 (MOT17) sequences and 42 simultaneous 30-FPS streams on an unspecified 16-gigabyte GPU. CNN-MV instead uses feed-forward updates and measures embedded latency, energy, and placement; stream capacity alone does not establish any of these. Our timing includes decoding, so no decoder-bypass benefit is claimed.

The two methods are alternatives: Analytical-MV favors latency and energy, whereas CNN-MV trades those advantages for higher recall and lower mean power. Host optimization accounts for most of the improvement over the original CNN-MV implementation; GPU placement then favors speed and energy, while DLA placement lowers GPU utilization. Scheduling must be chosen based on the surrounding workload, since parallelism improves DLA execution but not GPU execution in the evaluated configuration.

The present study is limited to 25 fixed-camera traffic videos. The different detector-routing rates are outcomes of each complete propagation-and-fallback policy; consequently, the evaluation captures system-level operating trade-offs but does not isolate propagation quality under a matched detector schedule. Moreover, Tegrastats provides power measurements at a coarser temporal resolution than the per-frame latency, and stage-level energy is therefore estimated through allocation rather than measured independently. The Nsight results provide system-level execution timelines rather than layer-level traces. Although the six CNN-MV training sequences are distinct from the 25 evaluation sequences, the internal training-validation split is sample-level rather than sequence-disjoint.

Future work will extend the evaluation to larger and more diverse datasets, including moving-camera scenarios, using sequence-disjoint training and validation partitions, and controlled detector schedules. Further investigation will consider adaptive refresh and fallback mechanisms, standard class-wise AP/mAP evaluation, and more detailed component-level energy profiling. The framework will also be extended to support workload-aware runtime scheduling that dynamically selects among Analytical-MV, CNN-MV, GPU execution, and DLA execution based on accuracy, latency, energy, and resource-contention requirements.

\section{Conclusion}
\label{sec:conclusion}

This paper presents two alternative compressed-domain propagation methods, evaluated using a common YOLO-based framework on a UA-DETRAC traffic-video subset and a Jetson AGX Orin platform. Analytical-MV jointly estimates translation and scale and is the overall efficiency leader in this evaluation. CNN-MV produces candidate boxes through learned residual updates, improves recall over Analytical-MV, and draws lower mean power than the non-CNN methods. GPU placement provides the fastest CNN-MV execution, while DLA placement reduces time-averaged GPU utilization. Across these configurations, optimized CPU batching, vectorization, and multithreaded scheduling show that host processing is as important as neural inference to end-to-end latency and energy. These results demonstrate the potential of motion-vector propagation for resource-constrained traffic-monitoring infrastructure and provide a basis for future evaluation in moving-camera physical-AI systems, including autonomous vehicles, drones, and mobile robots.

\vspace{-0.5em}
\ifCLASSOPTIONcaptionsoff
  \newpage
\fi

\bibliographystyle{IEEEtran}
\bibliography{Other/References}

\end{document}